\documentclass[10pt,twocolumn,letterpaper]{article}
\pdfoutput=1
\usepackage{cvpr}              

\usepackage[accsupp]{axessibility}

\usepackage{graphicx}
\usepackage{amsmath}
\usepackage{amssymb}
\usepackage{booktabs}

\usepackage{tabularx}
\newcolumntype{Y}{>{\centering\arraybackslash}X}

\usepackage{pifont}
\newcommand{\cmark}{\ding{51}}
\newcommand{\xmark}{\ding{55}}

\usepackage[dvipsnames]{xcolor}

\usepackage{xspace}

\makeatletter
\DeclareRobustCommand\onedot{\futurelet\@let@token\@onedot}
\def\@onedot{\ifx\@let@token.\else.\null\fi\xspace}

\def\eg{\emph{e.g}\onedot} 
\def\ie{\emph{i.e}\onedot}

\makeatother

\newcommand{\MainMethodAbbr}{GAIN}
\newcommand{\MainMethod}{Geometry-Aware Interaction Network}

\usepackage[pagebackref,breaklinks,colorlinks]{hyperref}

\usepackage[capitalize]{cleveref}
\crefname{section}{Sec.}{Secs.}
\Crefname{section}{Section}{Sections}
\Crefname{table}{Table}{Tables}
\crefname{table}{Tab.}{Tabs.}

\def\cvprPaperID{18} 
\def\confName{CVPR}
\def\confYear{2023}

\begin{document}

\title{A Closer Look at Geometric Temporal Dynamics for Face Anti-Spoofing}
\author{\begin{tabular}{c}
Chih-Jung Chang$^{12}$\thanks{Work done during the internship at Microsoft AI R\&D Center, Taiwan.}, Yaw-Chern Lee$^1$, Shih-Hsuan Yao$^1$, Min-Hung Chen$^1$, Chien-Yi Wang$^1$ \\ 
Shang-Hong Lai$^{13}$, Trista Pei-Chun Chen$^1$
\end{tabular} \\
$^1$Microsoft AI R\&D Center, Taiwan \quad $^2$Stanford University \quad
$^3$National Tsing Hua University, Taiwan
}
\maketitle

\begin{abstract}
Face anti-spoofing (FAS) is indispensable for a face recognition system. 
Many texture-driven countermeasures were developed against presentation attacks (PAs), but the performance against unseen domains or unseen spoofing types is still unsatisfactory. 
Instead of exhaustively collecting all the spoofing variations and making binary decisions of live/spoof, we offer a new perspective on the FAS task to distinguish between normal and abnormal movements of live and spoof presentations. We propose \textbf{\MainMethod~(\MainMethodAbbr)}, which exploits dense facial landmarks with spatio-temporal graph convolutional network (ST-GCN) to establish a more interpretable and modularized FAS model. Additionally, with our cross-attention feature interaction mechanism, \MainMethodAbbr~can be easily integrated with other existing methods to significantly boost performance.
Our approach achieves state-of-the-art performance in the standard intra- and cross-dataset evaluations. Moreover, our model outperforms state-of-the-art methods by a large margin in the cross-dataset cross-type protocol on CASIA-SURF 3DMask (+10.26 higher AUC score), exhibiting strong robustness against domain shifts and unseen spoofing types.
\end{abstract}

\begin{figure}[t]
\centering
\includegraphics[width=1.0\columnwidth]{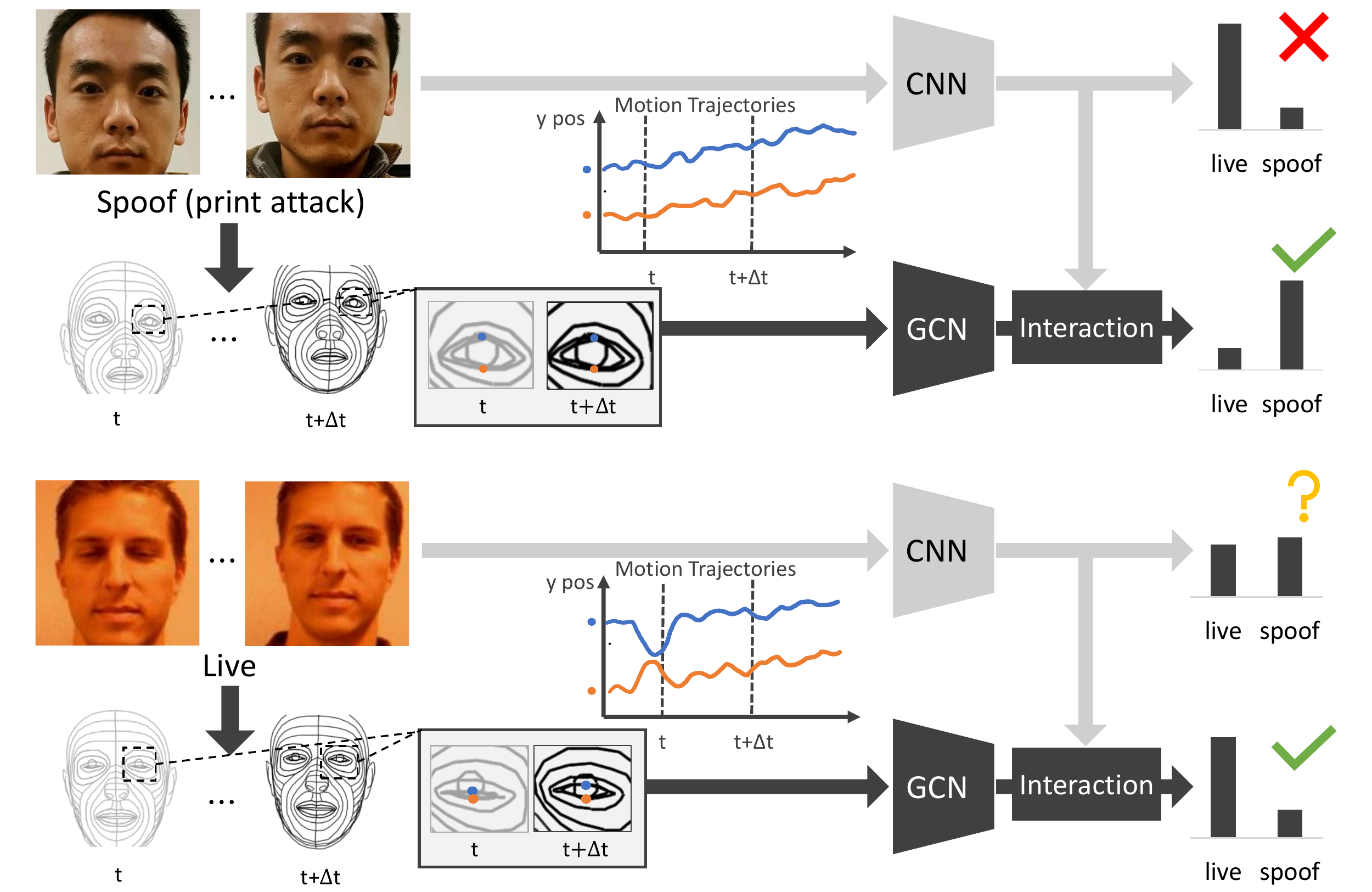}
\caption{Instead of only focusing on photometric information by common CNN-based FAS methods, our proposed \MainMethodAbbr~exploits geometric temporal dynamics based on dense facial landmarks to capture detailed facial motions, which is holistic and discriminative to enhance spoof detection. 
The motion trajectories correspond to the example facial landmarks, showing useful cues to distinguish live/spoof faces.}
\label{fig:teaser}
\end{figure}

\section{Introduction}

Face anti-spoofing (FAS) has played a critical role in securing face recognition systems from presentation attacks (\eg, print, video-replay, and 3D-mask attacks). Most popular FAS approaches extract fine-grained texture to spot spoofing cues~\cite{yang2014learn, atoum2017face, chen2019attention, yu2020searching, zhang2020face, yu2021dual}. These methods mainly adopt frame-level spoof detection, which aggregates the prediction of one or multiple frames to determine the liveness of a video instance. However, the extracted spoofing cues based on single-frame information might be insufficient to represent the characteristics of spoof faces. 

One common way to address the above issue is to exploit the \textit{temporal information} of videos. Live subjects differ from spoof attacks in motion (\ie, facial movements), indicating that dynamics in consecutive frames are beneficial to discriminating presentation attacks from live faces. CNN-LSTM/GRU networks~\cite{liu2018learning, yang2019face, saha2020domain, wang2020deep} and 3D-CNNs~\cite{xu2021exploiting, wang2021multi} have been used to extract temporal features from RGB frames. Despite the promising performance, focusing on RGB information may lead to overfitting to \textit{photometric} properties of live/spoof instances (\eg, material patterns) and under-exploring \textit{geometric} information (\eg, motion trajectories), which is important for robust temporal features (Fig.~\ref{fig:teaser}). Compared to a live subject's micro-movement, a paper-based attack commonly generates global motions such as translation or rotation, or abnormal transition movements of paper getting bent; for a 3D-mask attack, regardless of how realistic it appears, the movements introduced can be mostly treated as translation or rotation. All these motions appear relatively constrained in the geometric temporal modality, and some types of motions are even common across different attack types, indicating that \textit{geometric temporal information} is the key to further improving the model robustness against domain variations and unseen spoofing types. 
Therefore, the question becomes: \textit{How can we effectively exploit geometric temporal dynamics to benefit face anti-spoofing?}
Inspired by the success of skeleton-based action recognition~\cite{yan2018spatial, shi2019two, liu2020disentangling, chen2021channel}), which shows robustness to unseen domains by exploiting body joints, the use of \textit{landmarks} is motivated to effectively extract geometric temporal dynamics on faces.
In this work, we propose to improve the robustness of the temporal FAS features by extracting dynamic information on top of facial landmarks. Taking landmarks as inputs instead of raw RGB frames can explicitly extract geometric information that generalizes well to unseen attacking materials and is invariant under changes in cameras, lighting, and backgrounds. More specifically, we adopt dense facial landmark prediction~\cite{wood2022dense} to better capture the detailed discriminative motion from live faces and spoof attacks. In addition, we adopt a Graph Convolutional Network (GCN) to extract spatio-temporal features from dense facial landmarks, providing robust and representative dynamic information and reducing the computational complexity compared to the commonly used CNN-based temporal FAS methods. 
Finally, we design a cross-attention feature interaction strategy to integrate our geometric temporal features with photometric features from other FAS methods. The overall framework is dubbed \textit{\textbf{\MainMethod~(\MainMethodAbbr)}}.
Extensive experimental results on intra- and cross-dataset benchmarks demonstrate that the geometric feature of \MainMethodAbbr~is robust to unseen domains and provides liveness information complementary to current FAS methods, significantly boosting the performance.

Our contributions are three-fold:

\begin{itemize}
    \item To the best of our knowledge, we are the first to learn robust geometric temporal dynamics with the dense facial topology that captures fine-grained facial movements, which is critical to FAS but missing in previous works.
    \item In the proposed \MainMethodAbbr, our learned geometric temporal features can smoothly cooperate with photometric features from other existing FAS methods by our proposed cross-attention feature interaction strategy.
    \item The proposed \MainMethodAbbr~has been evaluated on intra-dataset, cross-dataset, and domain generalization benchmarks, achieving state-of-the-art performance and outperforming other FAS methods on all protocols. This demonstrates the efficacy of our proposed usage of geometric temporal dynamics.
\end{itemize}

\section{Related Works}

\paragraph{Face Anti-Spoofing:}
Most of the recent FAS methods aim at frame-level spoofing detection. They extract fine-grained information to identify spoofing patterns (\eg, lattice artifacts), using binary supervision~\cite{yang2014learn, chen2019attention} or leveraging auxiliary tasks~\cite{atoum2017face, yu2020searching, zhang2020face, yu2021dual}. In spite of promising results in intra-domain scenarios, these methods encounter degradation when evaluated in unseen domains, which motivates recent domain adaptation~\cite{li2018unsupervised, wang2019improving} and domain generalization~\cite{shao2019multi, jia2020single, wang2022domain, wang2022patchnet} approaches. Another group of FAS works can be viewed as temporal-based methods, which make use of temporal information in consecutive frames to extract liveness cues. These methods generally adopt CNN-LSTM/GRU networks~\cite{liu2018learning, yang2019face, saha2020domain, wang2020deep} or 3D-CNNs~\cite{xu2021exploiting, wang2021multi} to integrate spatial and temporal information. Nonetheless, by considering RGB information as input, such methods might run the risk of overfitting to photometric properties and neglecting geometric information that reveals facial motion patterns. In this work, we aim to extract robust geometric temporal features that focus on human facial dynamics to improve FAS performance.

\paragraph{Node-Based Action Recognition:}
In human action recognition, extracting dynamic information from key nodes (\ie, skeleton, and joint) trajectories has attracted much attention due to its robustness to scene variation. For example, sequences of human joints are fed into temporal CNNs~\cite{li2017skeleton, ke2017new, soo2017interpretable} or RNNs~\cite{shahroudy2016ntu, zhu2016co, liu2016spatio, zhang2017geometric} to predict human actions. However, these methods might overlook the inherent correlations of joints, thus having the limited capability of describing human dynamics. More recently, GCNs have demonstrated the power of leveraging graph topology to model joint correlations~\cite{yan2018spatial, shi2019two, liu2020disentangling, chen2021channel}. ST-GCN~\cite{yan2018spatial} applies GCNs along with temporal convolutions to learn both spatial and temporal information from landmark sequences, achieving impressive results in action recognition. Inspired by such success, we use GCN in this paper to extract geometric information on top of human facial landmarks and model facial movements.

\paragraph{Facial Landmark Prediction:}
Landmarks are instrumental in several face-related computer vision tasks. While sparse landmark prediction~\cite{bulat2017far, zhu2017face, guo2020towards} has been the mainstream approach, it fails to capture detailed facial characteristics and expressions since such information cannot be represented by a sparse set of landmarks. Recently, using dense landmarks has achieved remarkable results for 3D face reconstruction~\cite{wood2022dense}. By training with synthetic data, which guarantees dense yet consistent landmark annotations,\cite{wood2021fake} shows their dense landmark prediction model can produce accurate and expressive facial performance capture. The rich information provided by dense landmarks would be helpful to identifying fine-grained facial movements for spoof detection.

\begin{figure*}[t]
\centering
\includegraphics[width=0.97\textwidth]{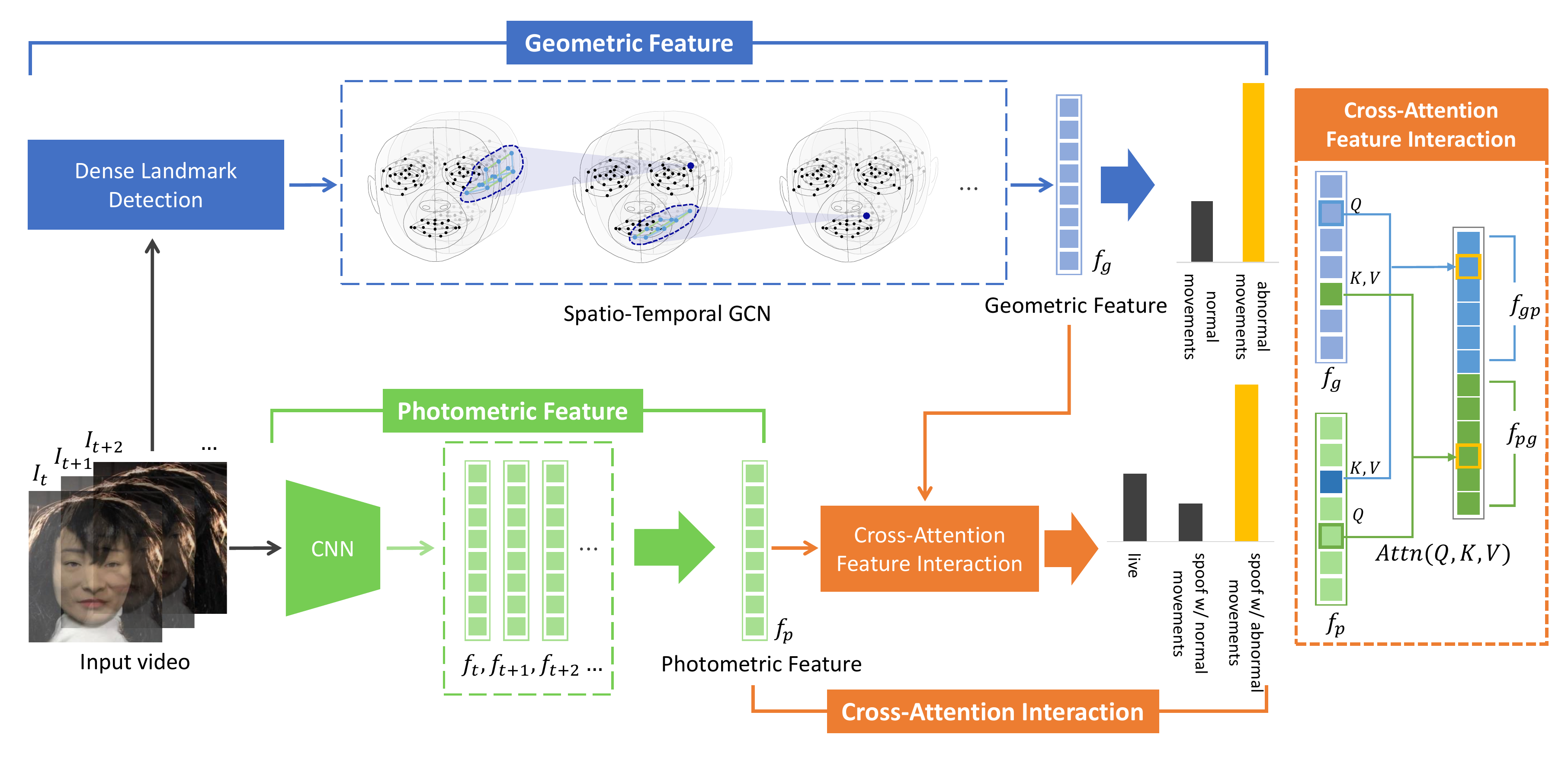}
\caption{The proposed \MainMethod. Aiming at exploiting robust temporal dynamics that benefit spoof detection, we learn geometric features $f_g$ (\textcolor{RoyalBlue}{blue}) by utilizing ST-GCN on top of dense landmark predictions. To integrate the learned $f_g$ with photometric features $f_p$ from common FAS methods (\textcolor{LimeGreen}{green}), we adopt a cross-attention mechanism (\textcolor{Orange}{orange}) that explores the interaction between geometric and photometric information, producing more discriminative features $f_{gp}$ and $f_{pg}$ for FAS.}
\label{fig:framework}
\end{figure*}

\section{Proposed Method}

With the aim of extracting robust temporal features to enhance FAS performance, we propose \MainMethod~(\MainMethodAbbr), which is designed to extract detailed facial dynamics from the geometric information, as illustrated in Fig.~\ref{fig:framework}. More specifically, we utilize GCN to extract features on top of dense facial landmarks and learn geometric information for fine-grained facial movements (Sec.~\ref{sec:dense_geometric_liveness_feature_learning}). We then integrate our learned geometric features with photometric features by the proposed cross-attention mechanism, further boosting the performance of SOTA FAS methods (Sec.~\ref{sec:cross-attention_feature_fusion}).

\subsection{Dense Geometric Liveness Feature Learning}
\label{sec:dense_geometric_liveness_feature_learning}

Facial movement is a discriminative component in describing live subjects. Formed by complex combinations of muscles, the detailed facial movements of live subjects are difficult to be produced by spoof attacks, such as print and multiple 3D-mask attacks, which usually produce \textit{abnormal} movements instead of \textit{normal} movements from live subjects. 
With the aim of robustly capturing such fine-grained movements, we propose to utilize GCN to learn geometric facial dynamics on top of dense facial landmarks.

\subsubsection{Dense Landmarks Detection}

In order to invariantly describe motion patterns in different domains, we exploit landmarks as representative geometric features. More specifically, we adopt dense landmark prediction~\cite{wood2022dense} to capture the fine-grained details of facial dynamics. As in~\cite{wood2022dense}, each facial landmark is predicted as a random variable $\{ v_i, \sigma_i \}$, where $v_i = (x_i, y_i)$ represents the expected position of a landmark, and $\sigma_i$ is a measure of uncertainty. The landmark prediction network is trained with synthetic data of densely annotated facial landmarks using a Gaussian negative log-likelihood (GNLL) loss:
\begin{equation}
    L_{GNLL}(v, \sigma) = \sum_{i=1}^{N} \lambda_i \left(\log({\sigma_i}^2) + \frac{{\lVert v_i - \hat{v_i} \rVert}^2}{2 {\sigma}^2} \right),
\end{equation}
where $N$ is the total number of facial landmarks, $\hat{v_i}$ is the training label for $v_i$, and $\lambda_i$ is the loss weighting.

We then utilize the trained network to predict dense facial landmarks for live and spoof subjects. Concretely, given an input RGB sequence $I = \{I_t\}_{t=1}^T$ of length $T$, we predict $N$ facial landmarks $\{v_{ti}\}_{i=1}^N$ for each frame $I_t$, where $v_{ti} = (x_{ti}, y_{ti})$ represents the expected position of a landmark. To even better spot the micro-movements on a face, we align the landmarks in each frame using simple yet effective frame-wise min-max normalization. Namely, we obtain aligned landmarks $v^\prime_{ti} = (x^\prime_{ti}, y^\prime_{ti})$ by:
\begin{equation}
    x^\prime_{ti} = \frac{x_{ti} - {x_t}_{min}}{{x_t}_{max} - {x_t}_{min}},
\end{equation}
where ${x_t}_{min}$ and ${x_t}_{max}$ denote $\min\limits_{i=1,\dots,N} x_{ti}$ and $\max\limits_{i=1,\dots,N} x_{ti}$ respectively, and $y^\prime_{ti}$ can be obtained likewise.
The visualization of facial landmarks is shown in Fig.~\ref{fig:ldmk}.

\subsubsection{Geometric Liveness Feature Learning}

With a rich set of facial landmarks obtained, we can learn the fine-grained motion of facial movements. Inspired by the success of GCN-based methods in skeleton-based human action recognition, we adopt the spatio-temporal graph convolution operator in ST-GCN~\cite{yan2018spatial} to model the 
geometrical relationships of facial landmarks across the frames. 

We first sub-sample $S$ frames out of $T$ aligned facial landmarks as the input of GCN. At each graph convolutional layer, a graph with $N \times S$ nodes is constructed, where each node is an input feature of size $C_{in}$. The input feature map can thus be represented by $f_{in} \in \mathbb{R}^{N \times S \times C_{in}}$. The edges of the graph are defined as the combination of: (1) the facial connectivity at each frame (Fig.~\ref{fig:ldmk}), and (2) the same facial landmark in consecutive frames. 
We represent the facial connectivity by the adjacency matrix $\textbf{A} \in \mathbb{R}^{N \times N}$, and additionally construct a learnable weight mask $\textbf{M} \in \mathbb{R}^{N \times N}$ for $\textbf{A}$ to adaptively learn edge weighting that helps capture detailed facial movements.
We formulate the higher-level feature map $f_{out}$ that integrates both spatial and temporal information as:
\begin{equation}
    f_{out} = \mathit{conv}(\textbf{$\Lambda$}^{-\frac{1}{2}} ((\textbf{A} + \textbf{I}) \odot \textbf{M}) \textbf{$\Lambda$}^{-\frac{1}{2}} f_{in} \textbf{W}),
\end{equation}
where $\textbf{I} \in \mathbb{R}^{N \times N}$ represents the identity matrix, $\textbf{$\Lambda$}^{ii} = \sum_j \textbf{M}^{ij} (\textbf{A}^{ij} + \textbf{I}^{ij})$ is for normalization, $\textbf{W} \in \mathbb{R}^{C_{in} \times C_{out}}$ is the weight for feature transformation, and $\odot$ and $\mathit{conv}(\cdot)$ denotes the element-wise product and a 1-D convolution along the temporal axis, respectively. 

After the last spatio-temporal convolutional layer, we apply global pooling across all $N \times S$ nodes to obtain the final feature $f_g$. To learn $f_g$ that discriminates between live facial dynamics and abnormal motions, we distinctly define the class of normal and abnormal movements: We group live faces and video-replay attacks together to form the class of normal facial movements, denoted as $G_{nm}$, and define the class of abnormal movements $G_{am}$ as attacks that cannot produce realistic motions (\eg, printed/displayed photos and plastic/plaster/resin masks). The objective of our geometric liveness learning $L_g$ is then formulated as:

\begin{equation}
    L_g = \mathit{BCE}(l_g, \textbf{W}f_g),
\end{equation}
where $\mathit{BCE}(\cdot)$ denotes the binary cross-entropy loss, $\textbf{W}$ represents a linear projection for classification, and $l_g$ is the label of the input sequence $I$, defined by:
\begin{equation}
    l_g = \begin{cases}
    1, & \text{if } I \in G_{nm}.\\
    0, & \text{if } I \in G_{am}.
    \end{cases}
\end{equation}

\subsection{Cross-Attention Feature Interaction}
\label{sec:cross-attention_feature_fusion}

Learned with the objective of identifying facial movements, the geometric feature obtained in Sec.~\ref{sec:dense_geometric_liveness_feature_learning} provides information complementary to common photometrics-based methods. To integrate the learned geometric features with these methods, it is important that the relationship between geometric and photometric features is highlighted and exploited. Fig.~\ref{fig:framework} illustrates our designed cross-attention strategy that fuses the learned geometric feature $f_g$ with feature extracted by common photometrics-based methods, $f_p$.

Given a chosen photometrics-based FAS network, we obtain the photometric feature $f_p$ by extracting the representation before the network's task-specific projection head. We then perform cross-attention on our geometric feature $f_g$ and the extracted $f_p$. Generally, the attention operation~\cite{vaswani2017attention} is computed as:
\begin{equation}
    \mathit{Attn}(Q, K, V) = \mathit{Softmax}(\frac{QK^\top}{\sqrt{d}})V,
\end{equation}
where $Q$, $K$, $V$ represent the input query, key, and value, and $d$ is the feature dimension. The interaction of $f_g$ and $f_p$ can thus be modeled by:
\begin{equation}
\begin{aligned}
    f_{gp} &= \mathit{Attn}(\textbf{W}_g^Q f_g, \textbf{W}_p^K f_p, \textbf{W}_p^V f_p), \\
    f_{pg} &= \mathit{Attn}(\textbf{W}_p^Q f_p, \textbf{W}_g^K f_g, \textbf{W}_g^V f_g),
\end{aligned}
\end{equation},
where each of the $\textbf{W}$s is a linear projection.

Finally, $f_{gp}$ and $f_{pg}$ are concatenated and fed into a linear classifier to predict the liveness probability. We train the cross-attention module using the cross-entropy loss, where live and spoof types are divided into 3 (instead of 2) classes: live faces, spoof attacks with normal facial movements, and spoof attacks with abnormal motions.
This encourages the geometric information of $f_g$ to be further exploited in the final decision and leads to more accurate spoof detection, as later verified in Sec.~\ref{sec:ablation_studies}. 

\section{Experiments}

\subsection{Datasets and Evaluation Metrics}

We evaluate the proposed \MainMethodAbbr~on five major public datasets: OULU-NPU~\cite{boulkenafet2017oulu}, CASIA-FASD~\cite{zhang2012face}, Replay-Attack~\cite{chingovska2012effectiveness}, MSU-MFSD~\cite{wen2015face}, and CASIA-SURF 3DMask~\cite{yu2020fas}. We utilize OULU-NPU for intra-dataset evaluation, and all five datasets for cross-dataset experiments. The performance is measured by the following evaluation metrics: Attack Presentation Classification Error Rate (APCER), Bona-fide Presentation Classification Error Rate (BPCER), Average Classification Error Rate (ACER), Half Total Error Rate (HTER), and Area Under Curve (AUC).
APCER, BPCER, and ACER are utilized for intra-dataset protocols, and HTER and AUC are used for cross-dataset evaluation.

\subsection{Implementation Details}
\begin{figure}[t]
\centering
\includegraphics[width=0.8\columnwidth]{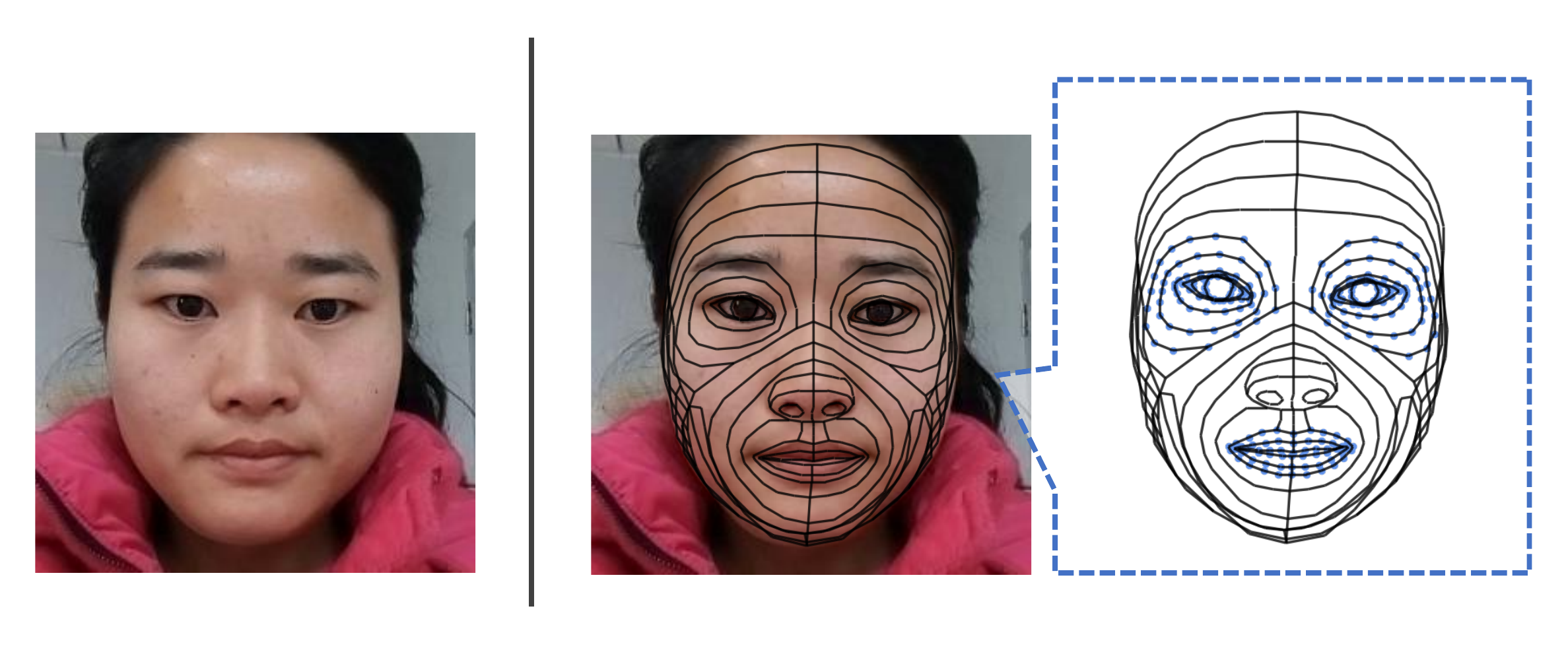}
\caption{Visualization of dense landmark prediction. The prediction consists of 550 landmarks, in which 246 landmarks around eyes and mouth region (visualized by the blue dots) are used in our proposed framework.}
\label{fig:ldmk}
\end{figure}

Given an input RGB sequence, we first detect and crop the face in each frame using RetinaFace~\cite{deng2020retinaface}, and then resize it to $256 \times 256$ before performing dense landmark prediction. A total of $N = 246$ landmarks around the eyes and mouth region (as shown in Fig.~\ref{fig:ldmk}) are chosen as the input of GCN. The sequence sub-sampling length $S$ is set to 64. Random sub-sampling is adopted during training. At the inference stage, we choose the 64 frames by selecting the landmarks with the highest variances of position from the input sequence. Mirror padding along the temporal axis is applied to any sequence of length less than 64.

The GCN is composed of 6 spatio-temporal graph convolutional units followed by a linear layer. We train the GCN for 65 epochs with a batch size of 16. The optimizer is SGD with a momentum of 0.9 and a weight decay of 0.0001. The initial learning rate is set to 0.1, and it is decayed with a factor of 0.1 after the 50-th epoch. We apply random rotation and horizontal flip to each landmark sequence for augmentation. For cross-attention feature interaction, we obtain $f_p$ using a chosen single-frame photometrics-based method by averaging features extracted from the frames in an input sequence. We specify the adopted photometrics-based method in each experiment in Sec.~\ref{sec:comparison_to_sota_methods}. The cross-attention module is trained for a maximum of 200 epochs with a batch size of 64. The optimizer is SGD with a momentum of 0.9, a learning rate of 0.1, and a weight decay of 0.0005.

\subsection{Comparison to SOTA Methods}
\label{sec:comparison_to_sota_methods}

\begin{table}[t]
    \begin{center}
    \scalebox{0.7}{\begin{tabularx}{1.47\columnwidth}{ll *3{Y}}
        \toprule
        Prot. & Method & APCER(\%) & BPCER(\%) & ACER(\%) \\
        \midrule
        & Auxiliary~\cite{liu2018learning} & 1.6 & 1.6 & 1.6 \\
        & ICLM~\cite{xu2021exploiting} & \underline{0.4} & \underline{0.0} & \underline{0.2} \\
        1 & NAS-FAS~\cite{yu2020fas} & \underline{0.4} & \underline{0.0} & \underline{0.2} \\
        & MTSS~\cite{huang2021multi} & 2.0 & 1.0 & 1.5 \\
        & CDCN++*~\cite{yu2020searching} & \textbf{0.0} & 0.8 & 0.4 \\
        & CDCN++* + \textbf{\MainMethodAbbr~(Ours)} & \textbf{0.0} & \textbf{0.0} & \textbf{0.0} \\
        \midrule
        & Auxiliary~\cite{liu2018learning} & 2.7 & 2.7 & 2.7 \\
        & ICLM~\cite{xu2021exploiting} & 2.5 & 1.9 & 1.6 \\
        2 & NAS-FAS~\cite{yu2020fas} & 1.5 & 0.8 & 1.2 \\
        & MTSS~\cite{huang2021multi} & \underline{1.4} & \textbf{0.3} & \underline{0.9} \\ 
        & CDCN++*~\cite{yu2020searching} & 1.9 & \underline{0.6} & 1.3 \\
        & CDCN++* + \textbf{\MainMethodAbbr~(Ours)} & \textbf{0.8} & 0.8 & \textbf{0.8} \\
        \midrule
        & Auxiliary~\cite{liu2018learning} & 2.7$\pm$1.3 & 3.1$\pm$1.7 & 2.9$\pm$1.5 \\
        & ICLM~\cite{xu2021exploiting} & 1.1$\pm$0.9 & 2.5$\pm$2.6 & 1.8$\pm$1.3 \\
        3 & NAS-FAS~\cite{yu2020fas} & 2.1$\pm$1.3 & 1.4$\pm$1.1 & 1.7$\pm$0.6 \\
        & MTSS~\cite{huang2021multi} & 2.1$\pm$1.3 & \textbf{0.5$\pm$0.4} & \underline{1.3$\pm$0.8} \\
        & CDCN++*~\cite{yu2020searching} & \textbf{0.6$\pm$0.5} & 2.8$\pm$4.2 & 1.7$\pm$2.0 \\
        & CDCN++* + \textbf{\MainMethodAbbr~(Ours)} & \underline{0.9$\pm$0.8} & \underline{0.8$\pm$0.8} & \textbf{0.9$\pm$0.8} \\
        \midrule
        & Auxiliary~\cite{liu2018learning} & 9.3$\pm$5.6 & 10.4$\pm$6.0 & 9.5$\pm$6.0 \\
        & ICLM~\cite{xu2021exploiting} & \textbf{3.3$\pm$3.9} & 4.1$\pm$4.4 & \underline{3.7$\pm$2.8} \\
        4 & NAS-FAS~\cite{yu2020fas} & \underline{4.2$\pm$5.3} & \textbf{1.7$\pm$2.6} & \textbf{2.9$\pm$2.8} \\
        & MTSS~\cite{huang2021multi} & 6.6$\pm$3.3 & \underline{2.4$\pm$2.8} & 4.5$\pm$2.2 \\
        & CDCN++*~\cite{yu2020searching} & 5.8$\pm$5.9 & 4.2$\pm$6.1 & 5.0$\pm$2.4 \\
        & CDCN++* + \textbf{\MainMethodAbbr~(Ours)} & 4.6$\pm$2.2 & 4.2$\pm$1.9 & 4.4$\pm$2.0 \\
        \bottomrule
    \end{tabularx}}
    \end{center}
    \caption{The results of the intra-dataset evaluation on OULU-NPU. We reproduce CDCN++ as our baseline method (noted as CDCN++*). The best results in each protocol are in \textbf{bold}, and the second-best ones are \underline{underlined}.}
    \label{tab:main_intra}
\end{table}


\subsubsection{Intra-Dataset Evaluation} 
For intra-dataset evaluation, we follow the protocols in OULU-NPU~\cite{boulkenafet2017oulu} to validate \MainMethodAbbr's generalization capacity under varied environments and presentation mediums. We adopt CDCN++~\cite{yu2020searching} as our baseline single-frame photometrics-based method. Table~\ref{tab:main_intra} compares our CDCN++-\MainMethodAbbr~fusion results with other FAS approaches, including those that also exploit temporal information. Combining learned geometric and extracted photometric features, \MainMethodAbbr~significantly improves the performance of CDCN++ in all the four protocols of OULU-NPU. Moreover, our results outperform other FAS methods in most settings, demonstrating the efficacy of learning geometric facial dynamics for FAS.

\begin{table*}[t]
    \begin{center}
    \scalebox{0.87}{\begin{tabularx}{1.1\textwidth}{l *8{Y}}
        \toprule
        & \multicolumn{2}{c}{O\&C\&I to M} & \multicolumn{2}{c}{O\&M\&I to C} & \multicolumn{2}{c}{O\&C\&M to I} & \multicolumn{2}{c}{I\&C\&M to O} \\
        \cmidrule(l){2-3} \cmidrule(l){4-5} \cmidrule(l){6-7} \cmidrule(l){8-9}
        Method & HTER(\%) & AUC(\%) & HTER(\%) & AUC(\%) & HTER(\%) & AUC(\%) & HTER(\%) & AUC(\%) \\
        \midrule
        Auxiliary (Depth)~\cite{liu2018learning} & 22.72 & 85.88 & 33.52 & 73.15 & 29.14 & 71.69 & 30.17 & 77.61 \\
        NAS-FAS~\cite{yu2020fas} & 19.53 & 88.63 & 16.54 & 90.18 & 14.51 & 93.84 & 13.80 & 93.43 \\
        ANRL~\cite{liu2021adaptive} & 10.83 & 96.75 & 17.85 & 89.26 & 16.03 & 91.04 & 15.67 & 91.90 \\
        DRDG~\cite{liu2021dual} & 12.43 & 95.81 & 19.05 & 88.79 & 15.56 & 91.79 & 15.63 & 91.75 \\ 
        SDFANet~\cite{zhou2021selective} & \underline{4.28} & 97.59 & 12.56 & 93.63 & \textbf{6.14} & 97.30 & 12.26 & 94.29 \\ 
        HFN+MP~\cite{cai2022learning} & 5.24 & 97.28 & 9.11 & \underline{96.09} & 15.35 & 90.67 & 12.40 & 94.26 \\ 
        VLAD-VSA (R)~\cite{wang2021vlad} & 4.29 & 98.25 & \underline{8.76} & 95.89 & 7.79 & \underline{97.79} & 12.64 & 94.00 \\ 
        SSAN-R~\cite{wang2022domain} & 6.67 & 98.75 & 10.00 & \textbf{96.67} & 8.88 & 96.79 & 13.72 & 93.63 \\ 
        AMEL~\cite{zhou2022adaptive} & 10.23 & 96.62 & 11.88 & 94.39 & 18.60 & 88.79 & \underline{11.31} & 93.96 \\ 
        LMFD-PAD~\cite{fang2022learnable} & 10.48 & 94.55 & 12.50 & 94.17 & 18.49 & 84.72 & 12.41 & 94.95 \\ 
        \midrule
        SSDG-R*~\cite{jia2020single} & 7.38 & 96.97 & 9.89 & 95.28 & 12.29 & 94.49 & 14.03 & 93.07 \\
        SSDG-R* + \textbf{\MainMethodAbbr~(Ours)} & \textbf{4.05} & \underline{98.92} & \textbf{8.52} & 96.02 & 8.50 & 97.27 & 12.50 & \underline{95.12} \\
        \midrule
        PatchNet*~\cite{wang2022patchnet} & 7.10 & 98.46 & 11.33 & 94.58 & 13.40 & 95.67 & 11.82 & 95.07 \\
        PatchNet* + \textbf{\MainMethodAbbr~(Ours)} & 4.29 & \textbf{99.45} & 10.67 & 95.13 & \underline{7.29} & \textbf{98.01} & \textbf{9.74} & \textbf{96.75} \\
        \bottomrule
    \end{tabularx}}
    \end{center}
    \caption{The results of the cross-dataset evaluation. We reproduce SSDG-R and PatchNet as our baseline methods (noted as SSDG-R* and PatchNet*). The best results in each protocol are in \textbf{bold}, and the second-best ones are \underline{underlined}.
    }
    \label{tab:main_cross}
\end{table*}
\begin{figure}[t]
\centering
\includegraphics[width=0.87\columnwidth]{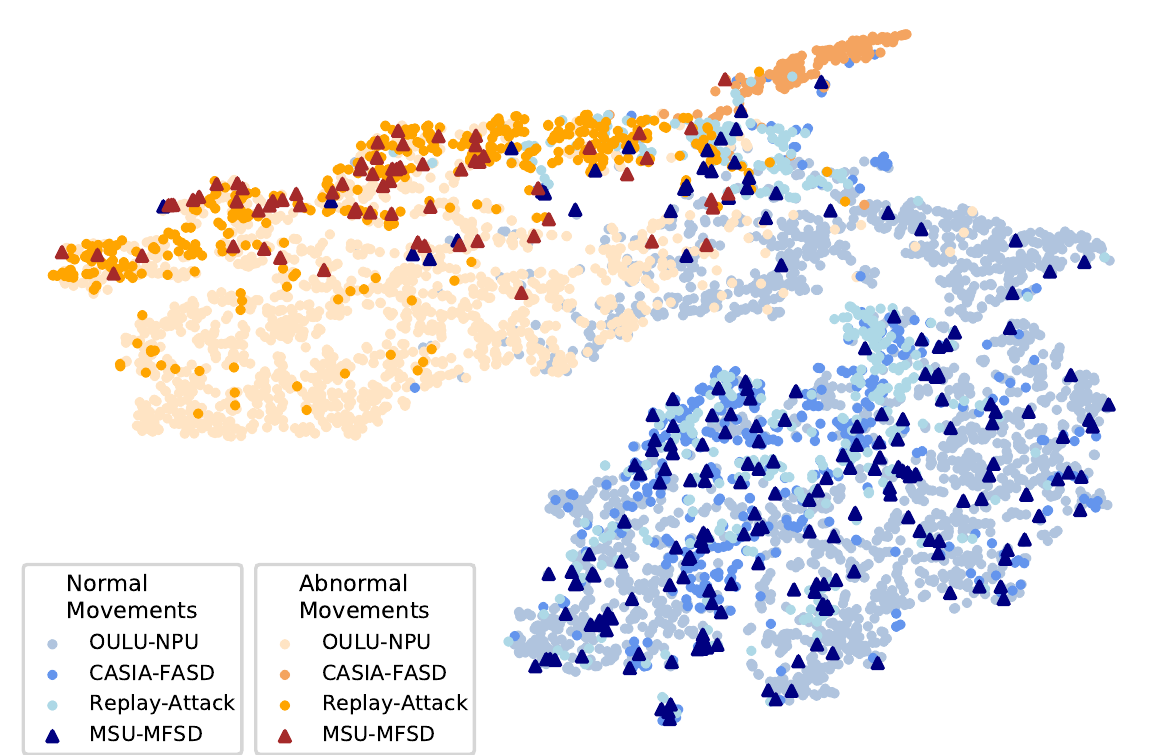}
\caption{t-SNE visualization of the extracted geometric features by \MainMethodAbbr~under the cross-dataset (O\&C\&I to M) setting. The distributions are well aligned between the testing dataset (M) and the training dataset (O\&C\&I).}
\label{fig:tsne_ocim}
\end{figure}

\subsubsection{Cross-Dataset Evaluation}
For cross-dataset evaluation, the following four datasets are used: OULU-NPU (denoted as O), CASIA-FASD (denoted as C), Replay-Attack (denoted as I), and MSU-MFSD (denoted as M). We follow the leave-one-out testing protocol in which one of the four datasets is selected for testing, and the remaining three are used for training (O\&C\&I to M, O\&M\&I to C, O\&C\&M to I, and I\&C\&M to O). The four datasets contain the same spoof types (live, print attack, and video-replay attack) while being collected under the variation of cameras, lighting conditions, resolutions, etc. This protocol indicates the models' generalization capability against unseen domains. 

We adopt two baseline single-frame methods, SSDG-R~\cite{jia2020single} and PatchNet~\cite{wang2022patchnet}, and show the results in Table~\ref{tab:main_cross}. \MainMethodAbbr~leads to the substantial performance boost over both of the baseline methods and reaches state-of-the-art results in the four settings. This verifies that facial motion patterns provide information complementary to the RGB-photometric modality, contributing to performance improvement. The results also demonstrate the strong generalization capability of our geometric features. As shown as well in Fig.~\ref{fig:tsne_ocim} by the t-SNE visualization~\cite{van2008visualizing}, a close match can be observed between the distributions of features extracted from the training (O\&C\&I) and testing (M) datasets.

\subsection{Ablation Studies}
\label{sec:ablation_studies}

\begin{table*}[t]
    \begin{center}
    \scalebox{0.87}{\begin{tabularx}{0.77\textwidth}{ccccc *2{Y}}
        \toprule
        & & & & & \multicolumn{2}{c}{O\&C\&I to M} \\
        \cmidrule(l){6-7}
        & Photometric & GCN Input & Attn & \# of Cls & HTER(\%) & AUC(\%) \\
        \midrule
        (a) & \cmark & \xmark & \xmark & 2 & 7.38 & 96.97 \\
        \midrule
        (b) & \xmark & dense & \xmark & 2 & 35.95 & 65.92 \\
        \midrule
        (c) & \cmark & dense & \xmark & 2 & 7.14 & 97.47 \\
        (d) & \cmark & dense & \xmark & 3 & 4.29 & 98.76 \\
        (e) & \cmark & dense & \cmark & 3 & \textbf{4.05} & \textbf{98.92} \\
        \midrule
        (f) & \cmark & sparse & \cmark & 3 & 5.71 & 98.12 \\
        \bottomrule
    \end{tabularx}}
    \end{center}
    \caption{Ablation study under the cross-dataset (O\&C\&I to M) setting. Note: \textit{Photometric} indicates whether the baseline photometrics-based method (SSDG-R) is adopted; \textit{GCN Input} specifies if the geometric features are learned from sparse or dense landmarks. \textit{Attn} indicates whether cross-attention is adopted in the method; \textit{\# of Cls} indicates if common live/spoof binary classification or the proposed 3-class classification is performed.}
    \label{tab:ablation}
\end{table*}
\begin{figure}[t]
\centering
\includegraphics[width=0.95\columnwidth]{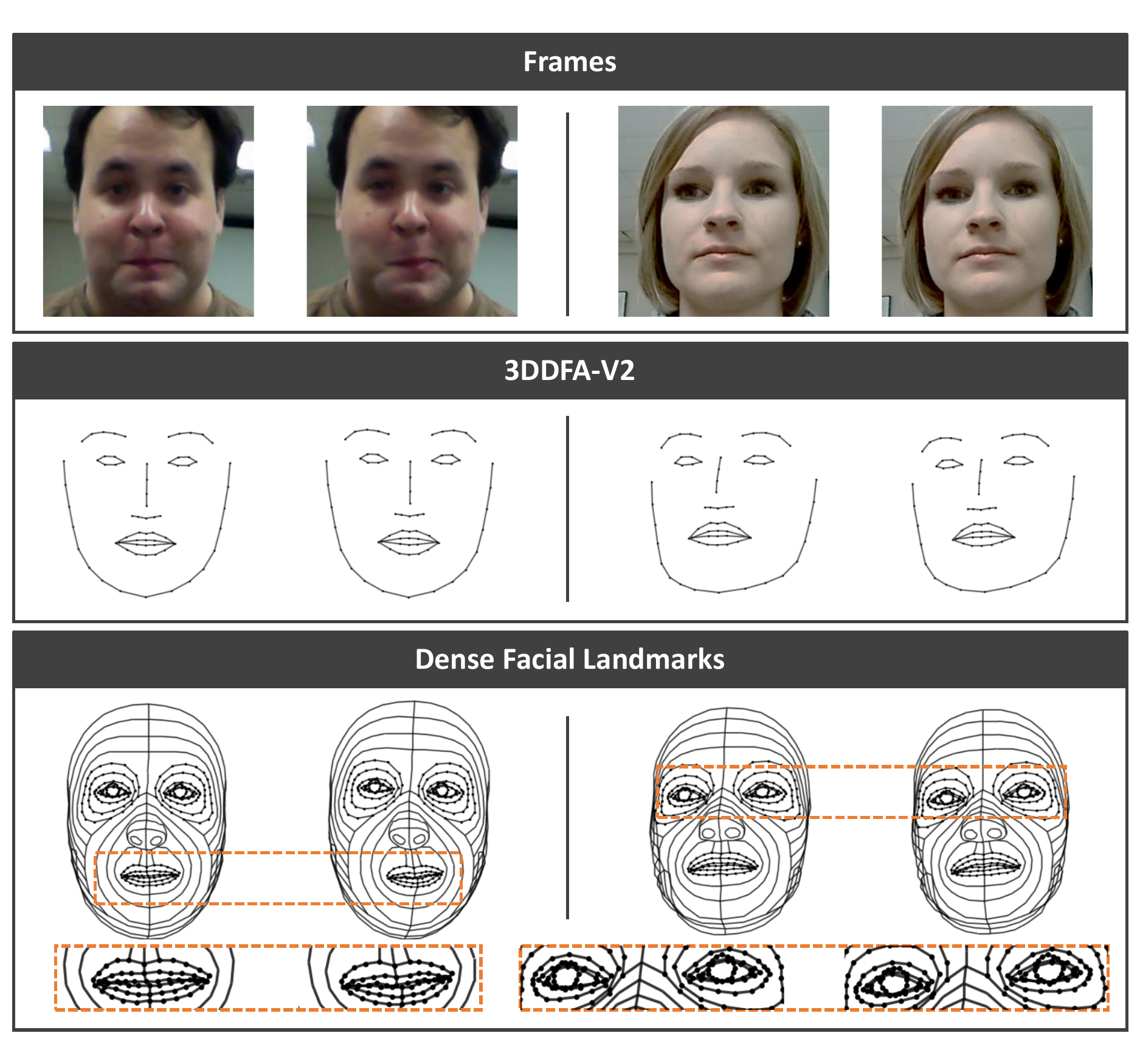}
\caption{Comparison of sparse and dense landmark predictions. The sparse landmarks (68 on each face) are predicted by 3DDFA-V2~\cite{guo2020towards}. Details such as pursing lips (left) and rolling eyes (right) are better captured by the dense landmarks.}
\label{fig:dense_sparse}
\end{figure}

\subsubsection{Benefit of Dense Landmark Prediction}

In our \MainMethodAbbr, dense landmark prediction is adopted to precisely capture fine-grained facial movements. In Fig.~\ref{fig:dense_sparse} we compare dense landmark predictions with sparse sets of 68 landmarks predicted by 3DDFA-V2~\cite{guo2020towards}. Dense landmarks show the superior ability to represent detailed movements such as pursing lips and rolling eyes. To quantitatively assess how using sparse and dense facial landmarks affect the learned geometric features, we integrate the features with the baseline photometrics-based method and report the performance in Table~\ref{tab:ablation} (e - f). Geometric features learned from dense landmarks lead to higher performance in spoof detection.

\begin{table}[t]
    \begin{center}
    \scalebox{0.87}{\begin{tabularx}{1.13\columnwidth}{l *2{Y}}
        \toprule
        \multicolumn{3}{c}{CASIA-SURF 3DMask}\\
        \midrule
        Method & HTER(\%) & AUC(\%)\\ 
        \midrule
        ResNet50~\cite{he2016deep} & 48.34 & 52.16 \\
        Auxiliary~\cite{liu2018learning} & 32.54 & 60.44 \\
        DTN~\cite{liu2019deep} & 38.97 & 69.24 \\
        NAS-FAS~\cite{yu2020fas} & 16.46 & 83.91 \\
        NAS-FAS w/ T(Mask)-Meta~\cite{yu2020fas} & 15.00 & 85.78 \\
        \midrule
        \textbf{\MainMethodAbbr~(ours)} & \textbf{6.23} & \textbf{96.04}\\
        \bottomrule
    \end{tabularx}}
    \end{center}
    \caption{The results of the cross-dataset cross-type evaluation. All methods are first trained on OULU-NPU with live, paper attacks, and replay attacks, and then evaluated on 3DMask dataset with unseen 3D-mask attacks.}
    \label{tab:gcn_3dmask}
\end{table}
\subsubsection{Analysis of Geometric Features}
It is important to note that the geometric features learned in our approach are not intended to address all possible liveness properties. Rather, their primary goal is to complement the common photometrics-based methods by providing critical temporal information. In Table~\ref{tab:ablation} (b), we present the results obtained by using geometric features alone for detecting spoof attacks. A significant drop in performance can be observed as the geometric features alone are insufficient to identify certain types of spoof attacks that can produce realistic facial movements, such as video-replay attacks. In Table~\ref{tab:ablation} (c), we combine geometric features with photometric features, enabling the model to leverage both sources of information for detecting spoof attacks. The result outperforms the photometrics-based baseline method (a).

While our learned geometric features alone may not be able to detect all types of spoof attacks, they are specifically suitable for spotting attacks that lack authentic facial dynamics and are quite robust. The CASIA-SURF 3DMask~\cite{yu2020fas} (or 3DMask) dataset, which does not include any attacks featuring realistic facial movements, is particularly 
well-suited for evaluating the performance of our proposed geometric features. 3DMask consists of realistic 3D mask attacks produced with 3D printing and is collected under various challenging lighting conditions (indoor/outdoor, with back/front/side-light). The realistic presenting materials and complex environments make it hard for conventional FAS solutions to discriminate between live and spoof subjects based on single-frame information.
We thus conduct the cross-dataset cross-type testing protocol proposed by the dataset to assess the ability of our geometric feature to detect unseen attacks with abnormal facial movements. Following the protocol, we first train \MainMethodAbbr~using OULU-NPU dataset with live, paper attacks, and replay attacks, and then evaluate the model on the unseen dataset (3DMask) with unseen 3D-mask attacks. As shown in Table~\ref{tab:gcn_3dmask}, our method significantly outperforms other photometrics-based FAS approaches. From this experiment, the robustness of our geometric feature learning is further justified.

\subsubsection{Comparison with Standard Temporal-Based Methods}

To exploit temporal information and focus on the \textit{geometric} instead of photometric properties, \MainMethodAbbr~utilizes GCN to extract geometric features on top of facial landmarks. In Table~\ref{tab:temporal_methods} we conduct two other experiments of abnormal movement detection, in which 3D-CNN (3D-ResNet-18~\cite{hara2018can}) is adopted as the backbone, as a comparison to showcase the strength of extracting geometric information from facial landmarks. The first row shows the method using 64 raw RGB frames of aligned faces. The input is of shape $(S, C, H, W) = (64, 3, 64, 64)$, where $S$ is the sequence sub-sampling length, $C$ is the number of input channels, and $H$ and $W$ are the height and the width of a frame. The second row utilizes the frame-wise information extracted by the standard optical flow algorithm~\cite{liu2009beyond} to generate the input of shape $(64, 2, 64, 64)$. Specifically, the two channels represent the orientation and magnitude of the calculated flow field. The experiment shows that Flow 3D-CNN and \MainMethodAbbr~provide significant improvements over the RGB 3D-CNN method, and \MainMethodAbbr~reaches the best performance. This indicates that pre-defined dense landmarks dedicated to a face can be more suitable for the FAS task compared to the generic-purpose optical flow. Moreover, the 
geometrical relationships of \MainMethodAbbr~bring stronger connectivity on the geometric and temporal domain.

\begin{table}[t]
    \begin{center}
    \scalebox{0.87}{\begin{tabularx}{0.93\columnwidth}{lY}
        \toprule
        \multicolumn{2}{c}{O\&C\&M to I} \\
        \multicolumn{2}{c}{Abnormal Movement Detection} \\
        \midrule
        Method & AUC(\%) \\ 
        \midrule
        RGB + 3D-CNN & 65.66 \\
        Flow + 3D-CNN & 92.28 \\
        \midrule
        \textbf{\MainMethodAbbr~(ours)} & \textbf{95.01} \\
        \bottomrule
    \end{tabularx}}
    \end{center}
    \caption{Comparison with standard temporal-based methods under the cross-dataset (O\&C\&M to I) setting. The performance is evaluated in terms of abnormal movement detection. \textit{RGB} and \textit{Flow} refer to using raw RGB frames and optical flow as the input, respectively.}
    \label{tab:temporal_methods}
\end{table}

\subsubsection{Feature Interaction Strategies}

We examine the effectiveness of each design in our feature interaction strategy in Table~\ref{tab:ablation} (c - e). The first row (c) shows the result of naively feeding concatenated features through linear layers and performing binary live/spoof classification. On the second row (d), by simply switching to the movement-aware 3-class classification (as described in Sec.~\ref{sec:cross-attention_feature_fusion}), the performance improves remarkably (-3.09\% in HTER and +1.26\% in AUC). It shows that facial dynamic information plays a critical role in FAS, and exploiting such information during feature integration results in improved spoof detection. Replacing linear layers with a cross-attention module (e) further encourages the model to explore the relationship between geometric and photometric features, leading to the full performance of \MainMethodAbbr.

\subsubsection{GCN Visualization}

\begin{figure}[]
\centering
\begin{subfigure}[]{1\columnwidth}
    \includegraphics[width=1.0\columnwidth]{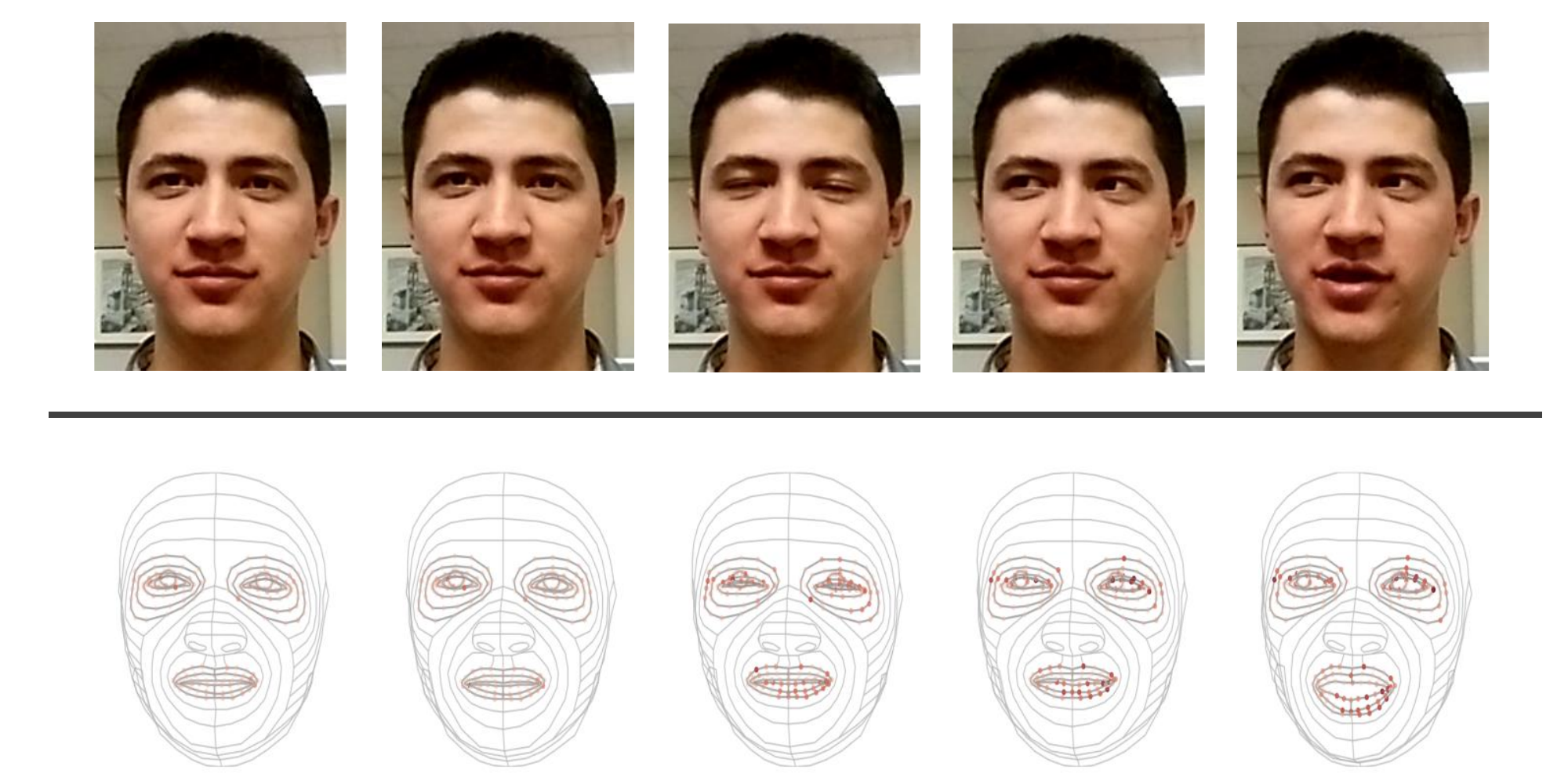}
    \caption{}
\end{subfigure}
\begin{subfigure}[]{1\columnwidth}
    \includegraphics[width=1.0\columnwidth]{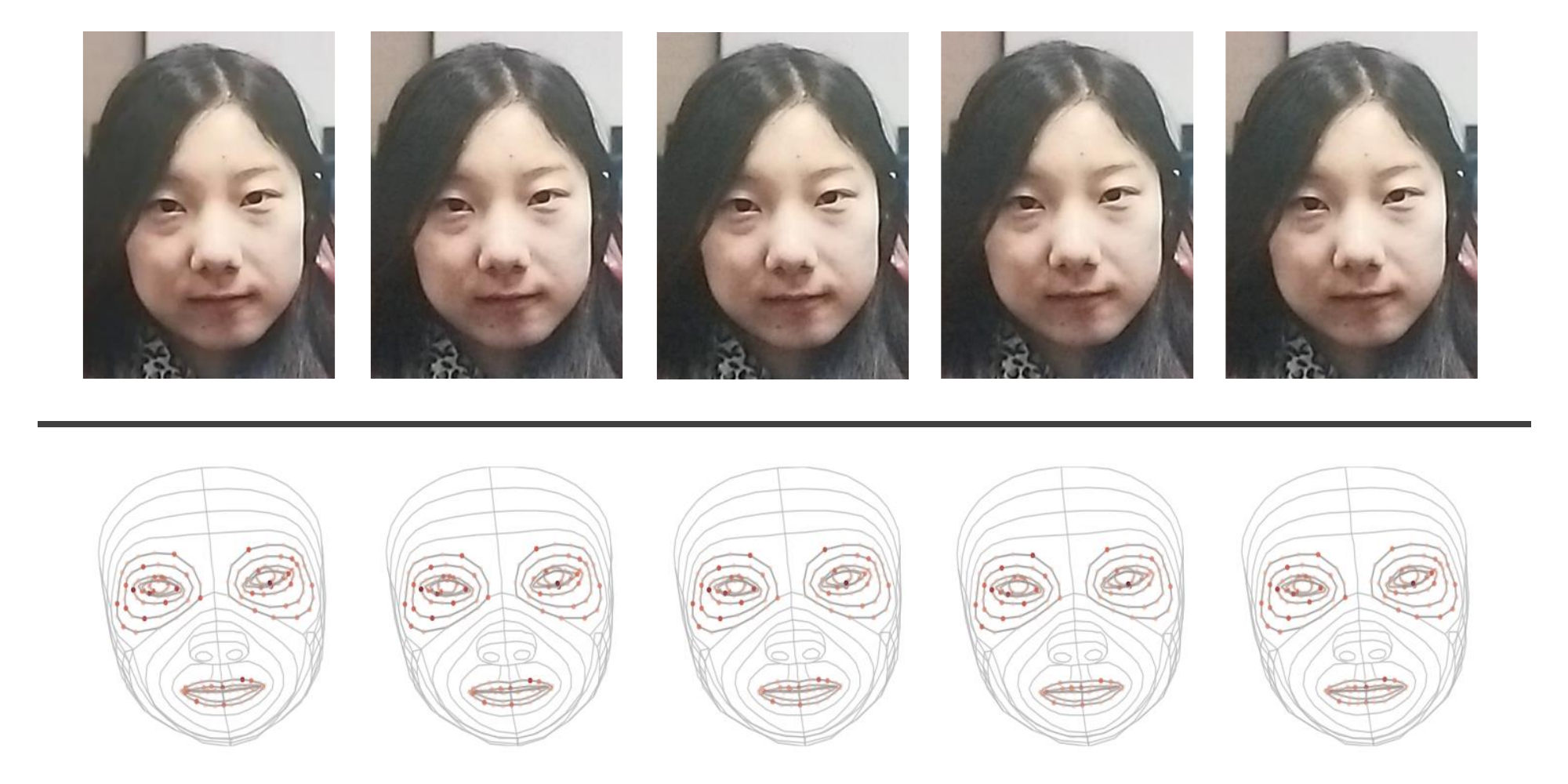}
    \caption{}
\end{subfigure}
\caption{GradCAM Visualization for GCN. The feature-node activations are derived from the last layer of the GCN along the temporal axis from (a) a live subject, and (b) a print attack. The activation intensities are indicated by the tones of red color.}
\label{fig:gradcamop2}
\end{figure}

Taking landmarks as inputs and training with the objective of movement classification, our GCN disentangles geometric temporal dynamics from texture information. In Fig.~\ref{fig:gradcamop2}, GradCAM for GCN~\cite{pope2019explainability} is implemented to interpret the geometric feature learned by the GCN. We visualize the feature-node activations along the temporal axis from the last graph convolutional layer. In live subjects, nodes tend to be more activated around the eye and mouth regions where facial motion can be observed at frames. Notably, activation intensities increase during movement transitions, such as raising eyebrows, moving eyeballs, and an open mouth. In contrast, similar intensities of attention on eye and mouth regions can be observed in the case of the print attack, in which the main motion is paper translation. The visualization once again verifies our proposed \MainMethodAbbr~is capable of capturing facial movements discriminatively.

\section{Conclusion}
In this work, we aim to learn geometric temporal dynamics that capture discriminative facial movements for face anti-spoofing. To achieve this goal, we propose \textit{\textbf{\MainMethod~(\MainMethodAbbr)}}, which adopts ST-GCN to extract fine-grained geometric facial dynamics from dense landmark predictions. The geometric features learned by \MainMethodAbbr~can be easily integrated with other photometrics-based methods using our designed cross-attention feature interaction strategy. Extensive experiments on diverse testing protocols (intra-, cross-dataset, and cross-type evaluations) demonstrate that the geometric features of \MainMethodAbbr~provide robust liveness information complementary to photometrics-based FAS methods, leading to a significant boost in performance.

{\small
\bibliographystyle{ieee_fullname}
\bibliography{egbib}
}

\end{document}